\documentclass[letterpaper]{article} 
\usepackage{aaai24}
\usepackage{times}  
\usepackage{helvet}  
\usepackage{courier}  
\usepackage[hyphens]{url}  
\usepackage{graphicx} 
\usepackage{bm}
\usepackage{amsmath}
\usepackage{amsfonts}
\usepackage{natbib}  
\usepackage{caption} 
\usepackage{algorithm}
\usepackage{algorithmic}

\newtheorem{theorem}{Theorem}[section]

\newtheorem{definition}[theorem]{Definition}

\newcommand{\vo}{\boldsymbol{o}}
\newcommand{\name}{NER}
\usepackage{newfloat}
\usepackage{listings}
\DeclareCaptionStyle{ruled}{labelfont=normalfont,labelsep=colon,strut=off} 
\floatstyle{ruled}
\newfloat{listing}{tb}{lst}{}
\floatname{listing}{Listing}
\title{Never Explore Repeatedly in Multi-Agent Reinforcement Learning}
\author{
    Chenghao Li\textsuperscript{\rm 1},
    Tonghan Wang\textsuperscript{\rm 2},
    Chongjie Zhang\textsuperscript{\rm 3},
    Qianchuan Zhao\textsuperscript{\rm 1}
}
\affiliations{
    \textsuperscript{\rm 1}Tsinghua University, \textsuperscript{\rm 2}Harvard University, \textsuperscript{\rm 3}Washington University (St. Louis)\\

    $\{$lch.tsinghua, tonghanwang1996, zhangchongjie$\}$@gmail.com, zhaoqc@tsinghua.edu.cn
}

\usepackage{bibentry}

\begin{document}

\maketitle

\begin{abstract}
In the realm of multi-agent reinforcement learning, intrinsic motivations have emerged as a pivotal tool for exploration. While the computation of many intrinsic rewards relies on estimating variational posteriors using neural network approximators, a notable challenge has surfaced due to the limited expressive capability of these neural statistics approximators. We pinpoint this challenge as the "revisitation" issue, where agents recurrently explore confined areas of the task space. To combat this, we propose a dynamic reward scaling approach. This method is crafted to stabilize the significant fluctuations in intrinsic rewards in previously explored areas and promote broader exploration, effectively curbing the revisitation phenomenon. Our experimental findings underscore the efficacy of our approach, showcasing enhanced performance in demanding environments like Google Research Football and StarCraft II micromanagement tasks, especially in sparse reward settings.
\end{abstract}

\section{Introduction}

To introduce intelligence into multi-agent systems~\citep{zhang2013coordinating,zhang2019integrating} and achieve sophisticated cooperative behavior, multi-agent reinforcement learning (MARL) has been gaining increasing interest in recent years. Advanced MARL methods have significantly pushed forward the performance of machine learning algorithms on tasks such as the micromanagement of units in StarCraft II~\citep{rashid2018qmix, wang2021rode}, the strategic card game Hanabi~\citep{bard2020hanabi,foerster2019bayesian}, and the control of robotic systems~\citep{kurin2020my}. 

Nevertheless, a persistent bottleneck hampers the full realization of MARL's potential in more intricate and complex problem domains. The action-observation space grows exponentially with the number of agents, posing a challenge to the exploration efficiency within the vast search space, subsequently affecting the overall learning efficiency of MARL algorithms. Basic exploration strategies, such as the widely-used $\epsilon$-greedy approach~\citep{wang2020qplex,yu2021surprising,de2020independent}, appear to be inadequate, faltering even in environments with a moderate number of agents.

Existing methods for multi-agent exploration primarily utilize intrinsic rewards to enhance exploration. These rewards are typically designed to stimulate either individual or collective agent behavior, often characterized by specific behavioral statistics. For example, \citet{wang2020influence} encourages inter-agent interaction, depending on estimating the probability $p(s_i'| s_i, a_i)$, where $s_i$ and $a_i$ denote the local state and action of agent $i$, respectively. \citet{houthooft2016vime} aims to maximize measurements of behavioral randomness, relying on $p(\theta)$ where $\theta$ represents policy parameters. \citet{jiang2021emergence} promotes individuality by estimating $p(i|o_i)$, and \citet{li2021celebrating} advocates for behavioral diversity by estimating $p(o_i'|\tau_i,a_i,i)$, with $o_i$ and $\tau_i$ symbolizing agent $i$'s observation and action-observation history, respectively. However, the statistical estimation of these distributions often proves intractable in continuous learning tasks. To overcome this challenge, previous methods have integrated variational inference techniques~\citep{alemi2017deep} into multi-agent settings, employing neural network approximators to estimate the variational posterior. 

\begin{figure*}[t]
\centering
\includegraphics[width=\linewidth]{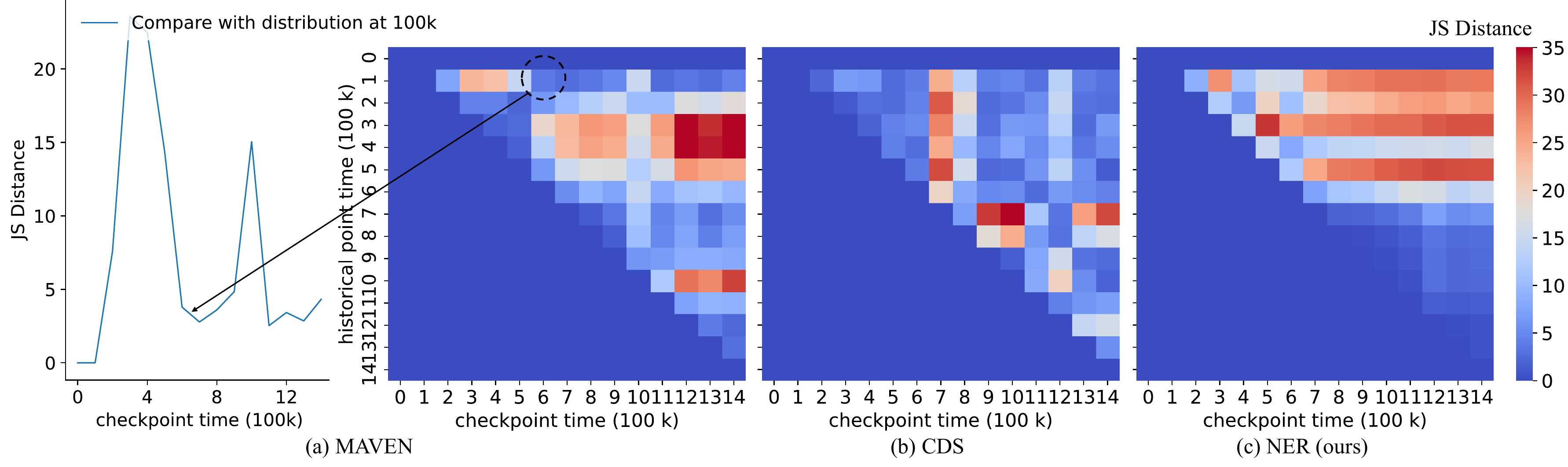}
\vspace{-2em}
\caption{The issue of revisitation measured with Jensen-Shannon (JS) distances. Experiments are carried out on a 6$\times$12 maze task shown in Fig.~\ref{fig:env_maze}. During learning, we store joint observation distributions every 100$k$ time steps, named historical points. Meanwhile, for every 100k time step, we calculate the JS distance between the current distribution (checkpoint) and all historical points. \textbf{(a)} JS distances from historical points of MAVEN~\citep{mahajan2019maven}. We observe the revisitation issue as the joint observations explored at 100$k$ are similar to those explored after 500$k$. \textbf{(b)} CDS~\citep{li2021celebrating} also suffers from the revisitation issue. Only the joint observations explored at 700$k$ are significantly different from those at other time steps. \textbf{(c)} Our method avoid revisitation by dynamically scaling the intrinsic rewards of CDS. The Jensen-Shannon (JS) distances exhibit a tendency to increase until convergence.}
\label{fig:teaser}
\end{figure*}

While variational approximation offers a practical solution to the calculation of various intrinsic rewards, which has spurred significant advancements in MARL, enabling enhanced exploration and addressing complex tasks such as Google Research Football (GRF) academy scenarios~\cite{kurach2020google}, in this paper, we argue that the limited generalization ability of variational approximators limits estimation accuracy and leads to a problem which we call \textbf{\emph{revisitation}}. This term refers to the situation where agents forget what they have visited before after the detachment~\citep{ecoffet2019go} of exploration, causing them to re-explore with a shrinking exploration boundary. Such frequent revisitation not only results in a wasteful training process characterized by a substantial accumulation of redundant experiences, but may also leave a large action-observation sub-space unexplored even after a prolonged training process.

To elucidate this concept, we analyze the joint observation distribution visited by MAVEN~\citep{mahajan2019maven} and CDS~\citep{li2021celebrating} in a 6$\times$12 maze (Fig.~\ref{fig:env_maze}), as illustrated in Fig.\ref{fig:teaser}. As measured by the Jensen-Shannon distance~\citep{endres2003new} between distributions, MAVEN exhibits a tendency to re-explore the joint observation visited at around the 100$k$ timestep, after 500$k$ training timesteps. This pattern of periodic revisitation is not unique to MAVEN; a similar trend is also observed for CDS.

In this paper, we begin by formally defining the revisitation issue, a challenge that has significant implications for the efficiency of MARL algorithms. To address this problem, we introduce an intrinsic reward dynamic scaling approach, specifically tailored to mitigate the effects of revisitation. The core principle of our method involves modulating intrinsic rewards based on the familiarity of joint observation distributions. This modulation helps offset the substantial variances caused by the estimation errors of variational approximators, while encouraging the coverage of the joint observation space.

Our proposed solution, named NER (\textbf{N}ever \textbf{E}xplore \textbf{R}epeatedly), effectively reduces revisitation and fosters continuous exploration. As illustrated in Fig.~\ref{fig:teaser} (right), within the maze task, the Jensen-Shannon (JS) distances between historical data points consistently increase throughout the learning period until convergence. Furthermore, we validate our approach by benchmarking it on both Google Research Football \cite{kurach2020google} and StarCraft II Micromanagement \cite{samvelyan2019starcraft} tasks under sparse reward settings. The results are promising: NER not only effectively manages the issue of revisitation but also advances the state-of-the-art in the field. This paper, therefore, presents a novel perspective and offers a practical solution to the complex problem of revisitation, laying the groundwork for further innovation and exploration in the field of multi-agent reinforcement learning.

\section{Related Work}

In this paper, we identify and concentrate on the revisitation problem that occurs in the multi-agent setting. Here we introduce the related work.

\textbf{Exploration Bonuses.} 
Exploration is a pivotal factor that significantly influences the efficiency of deep reinforcement learning~\citep{houthooft2016vime,trott2019keeping, raileanu2020ride, zintgraf2021exploration, chen2021contingency}. Commencing with the classic $\epsilon$-greedy action selection strategy, a plethora of advanced algorithms have been formulated to amplify the efficacy of exploration. Count-based exploration techniques, for instance, employ visit counts to steer an agent's behavior, aiming to minimize uncertainty~\citep{strehl2008analysis, bellemare2016unifying, martin2017count, ostrovski2017count, tang2017exploration, machado2020count}. One illustrative method is Random Network Distillation (RND), which utilizes prediction errors to compute count-based rewards, thereby adapting to a continuous state space~\citep{burda2018exploration, osband2018randomized, osband2019deep, ciosek2019conservative}. This principle of leveraging prediction errors has also been employed in works like \cite{pathak2017curiosity, burda2018large, pathak2019self, dean2020see} to foster self-organized intrinsic motivations, thereby contributing to the richness and robustness of exploration techniques in reinforcement learning.

\textbf{Exploration in the Multi-Agent Setting.} Exploration becomes even more vital in the multi-agent setting, where the search space expands exponentially with the number of agents. In addition to probing the environment, it's necessary to foster diversity across agents to facilitate complex coordination. Within the multi-agent context, agents' exploration may be mutually encouraged~\citep{mahajan2019maven,gupta2021uneven,zheng2021episodic,liu2021cooperative}, promoted in pairs~\citep{wang2020influence,ndousse2021emergent}, or individually incentivized to stand out from the whole group~\citep{jiang2021emergence, li2021celebrating}. Exploration can further be stimulated through the segregation of roles~\citep{wang2020roma, wang2021rode}. In this paper, we specifically select CDS~\citep{li2021celebrating} as the intrinsic motivation for exploration, as it represents the current state-of-the-art algorithm across various challenging benchmarks.

\textbf{Forgetting.} Revisitation, as explored in our paper, is intrinsically linked to the phenomenon of forgetting past policies. This concept has been extensively studied in the realm of lifelong learning, particularly in multi-task settings~\citep{rebuffi2017icarl, li2017learning, rolnick2019experience, von2019continual}. Several cutting-edge algorithms have delved into this issue, adopting approaches like supervised regularization~\citep{kirkpatrick2017overcoming} or replay mechanisms~\citep{isele2018selective, yan2022learning}, further amplifying the stability-plasticity conundrum. Some have even proposed multi-stage processes, where an agent first leverages existing knowledge for the current task and then assimilates new insights into a communal knowledge base~\citep{schwarz2018progress,mendez2020lifelong}. Yet, the notion of forgetting remains relatively uncharted in the context of single-task, multi-agent reinforcement learning. This oversight is surprising, given the inherent potential for agents to forget during exploration phases. The crux of the matter lies in devising methods to automatically stave off this forgetting, without granular insights into the agents' progression during learning. This challenge demands meticulous scrutiny and inventive solutions. Our research endeavors to make a meaningful contribution to this burgeoning area of inquiry.
\section{Method}\label{sec:method}

In this section, we first provide a formal definition of the concept of revisitation within the context of multi-agent settings. Following that, we detail our approach to dynamically scaling intrinsic rewards, specifically designed to address and mitigate the issue of revisitation.

We consider multi-agent cooperation tasks that can be modeled as Dec-POMDP~\cite{oliehoek2016concise} $\mathcal{G}=\langle I, S, A, P, R, \Omega, n, \gamma\rangle$, where $I$ is the set of agents, $S$ is the state space, $A$ is the action space, $P$ is the transition function, $R$ is the reward function, $O$ is the observation space, $n$ is the number of agents, and $\gamma \in [0, 1)$ is the discount factor. At each time step, each agent $i \in I$ receives its local observation $o_i\in \Omega$ and selects an action $a_i \in A$ based on its action-observation history $\tau_{i} \in \mathrm{T} \equiv(\Omega \times A)^{*} \times \Omega$. According to the joint action $\bm{a}$ and environment transition function $P\left(s^{\prime} \mid s, \bm{a}\right)$, the environment transfers to a new state $s'$ and provides an environment reward $r^e=R(s, \bm{a})$ that is shared across all agents. Previous multi-agent exploration methods additionally associate an intrinsic reward $r^I$ with each transition. In this work, we study how to dynamically adjust $r^I$ to avoid revisitation. We call $\mathcal{G}_p=\langle I, S_p, A, P, R, \Omega_p, n, \gamma\rangle$ a \emph{sub-space} of $\mathcal{G}$ where $S_p\subset S$ and $\Omega_p\subset \Omega$.

\subsection{Revisitation Characterization}\label{sec:method-detect}
As previously discussed, revisitation refers to the scenario in which agents explore samples that have been previously explored. In multi-agent settings, we define the concept of revisitation with reference to the joint observation distribution, which is induced by the collective policy $\boldsymbol{\pi}$ employed by the agents: $\rho_{\boldsymbol{\pi}}(\boldsymbol{o})=\mathbb{E}_{\boldsymbol{\pi}}\left[\frac{1}{T} \sum_{t=0}^{T} \mathbf{1}_{\boldsymbol{o}_{t}=\boldsymbol{o}}\right]$. Intuitively, revisitation happens, if after a significant change in joint observation distribution, $\rho_{\boldsymbol{\pi}}(\boldsymbol{o})$ degenerates to a distribution that has occurred before. Formally, we have the following definition:

\begin{definition}
A revisitation occurs at time step $t$ if there exists a positive constant $\delta$ and a time step $t' < t$ such that 
\begin{equation}
\begin{aligned} \sup _{\tau \in\left(t^{\prime}, t\right)} & D_{\mathrm{JS}}\left[\rho_{\boldsymbol{\pi}_{\tau}}(\boldsymbol{o}) \| \rho_{\boldsymbol{\pi}_{t}}(\boldsymbol{o})\right]>\delta \\ & D_{\mathrm{JS}}\left[\rho_{\boldsymbol{\pi}_{t^{\prime}}}(\boldsymbol{o}) \| \rho_{\boldsymbol{\pi}_{t}}(\boldsymbol{o})\right]<\delta\end{aligned}
\label{def:revisitation}
\end{equation}
\end{definition}

Here, $D_{\mathtt{JS}}$ represents the Jensen-Shannon distance between two distributions. The first condition characterizes the capability of many multi-agent exploration algorithms. As learning progresses, agents are encouraged to explore new action-observation sub-spaces, leading to a shift in the joint observation distribution. The second condition, on the other hand, illustrates the phenomenon of revisitation, signifying that agents are re-exploring the sub-spaces they have previously visited. This dual-condition framework helps us better understand the dynamics of exploration and provides insights into the occurrence of revisitation, which our proposed methods aim to address.

\begin{figure}[t]
\centering
\includegraphics[width=\linewidth]{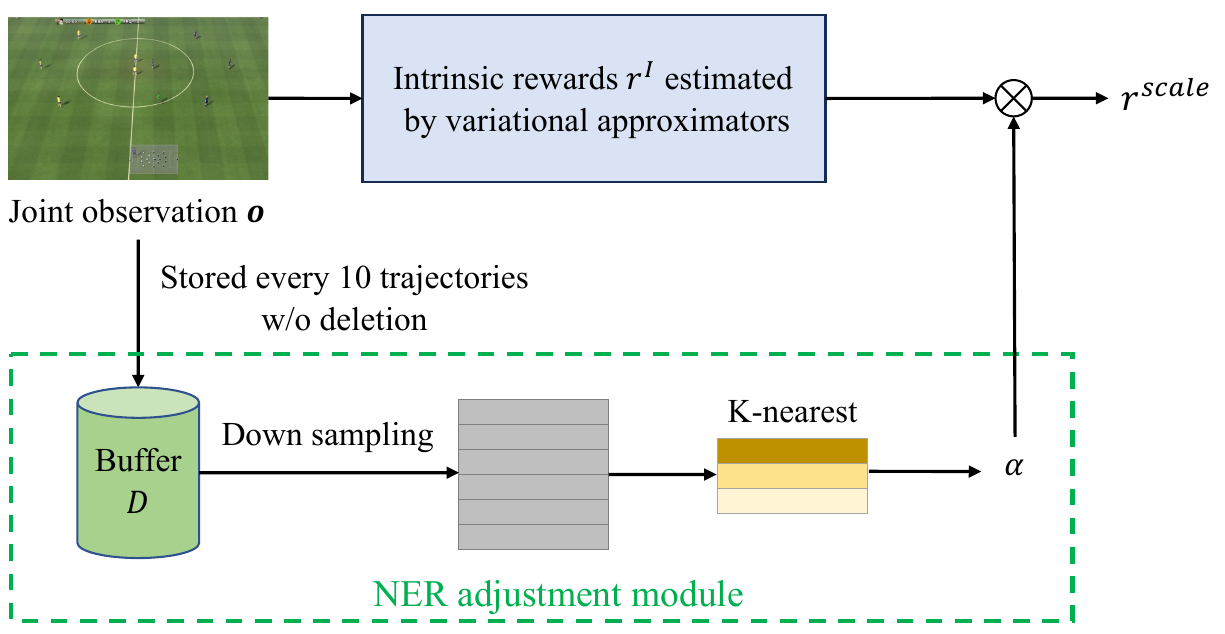}
\vspace{-2em}
\caption{NER (\textbf{N}ever \textbf{E}xplore \textbf{R}epeatedly)'s intrinsic reward dynamic adjustment module.}
\label{fig:reward}
\end{figure}

\begin{figure*}[t]
\centering
\includegraphics[width=1.\linewidth]{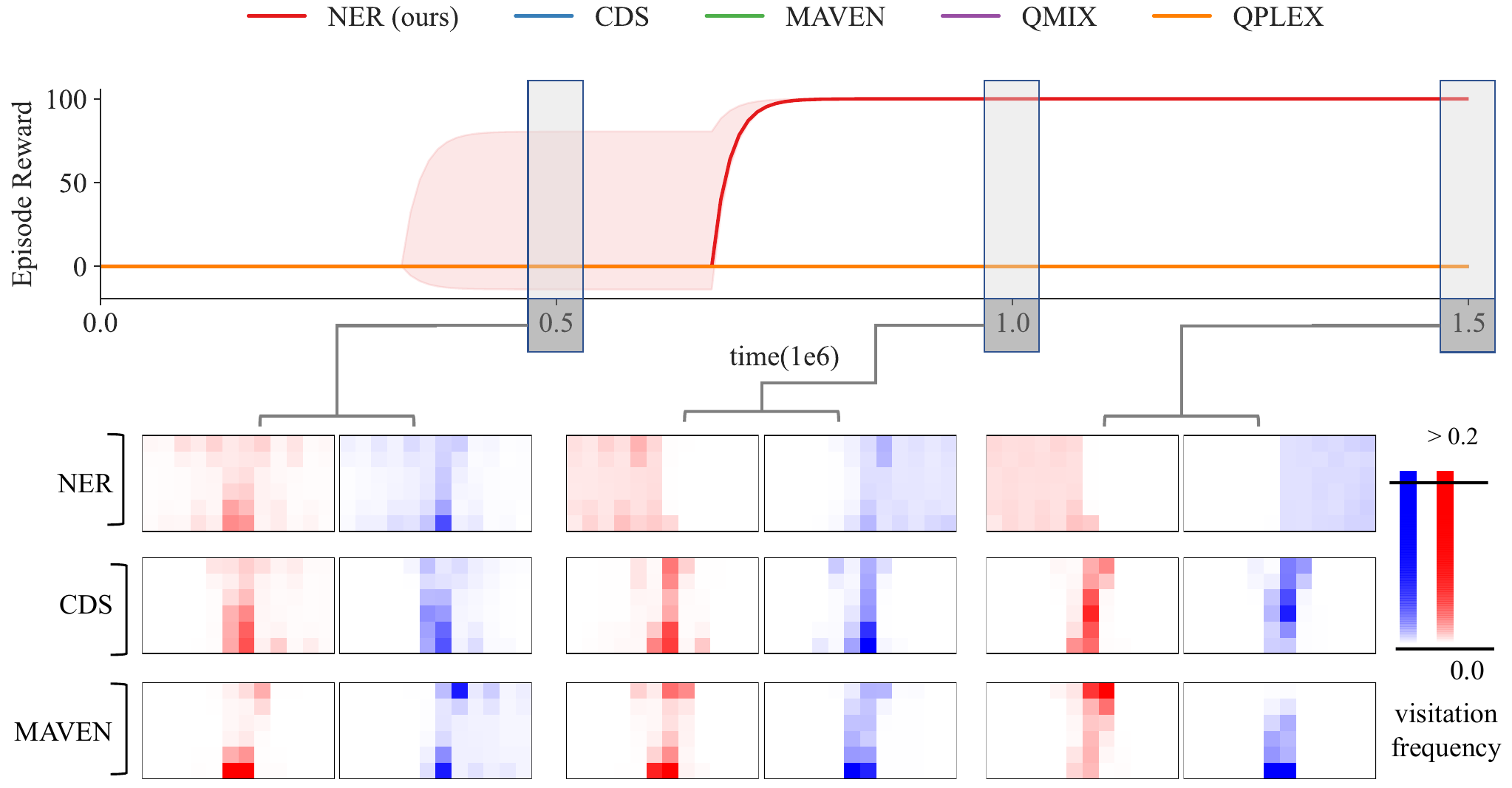}
\caption{Comparison against baselines on the $\mathtt{Maze}$ environment with respect to learning performance (Up) and visitation heat maps during training (Down). Current intrinsic motivations, such as MAVEN and CDS that utilize variational approximators, tend to repetitively explore areas close to their starting points, leading to a diminishing exploration boundary. In contrast, our approach circumvents this revisitation issue and attains superior exploration outcomes.}
\label{fig:maze_result}
\end{figure*}

\subsection{Dynamic Intrinsic Reward Scaling}\label{sec:method-reward}

To address the issue of revisitation and achieve efficient exploration, we propose a method to dynamically scale the intrinsic rewards $r^I$ estimated by variational approximators:
\begin{align}
	r^{\mathtt{scale}} = \alpha r^I
\label{equ:all_reward}.
\end{align}

We posit that the scaling factor $\alpha$ should be minimized in well-explored sub-spaces. This reduction aims to counterbalance the pronounced variance of $r^I$, which arises due to the increased error of its neural-network-based variational approximators after exploring other sub-spaces. In contrast, within less familiar sub-spaces, $\alpha$ should be large to stimulate further exploration and encourage the coverage of the entire action-observation space.

To avoid introducing additional estimation uncertainty when calculating $\alpha$, we consciously refrain from using neural networks in its determination. Specifically, for the intrinsic reward associated with the transition $(s,\bm o,\bm a,s',r^e,\vo')$, we adjust $\alpha$ according to the similarity of $\vo$ to joint observations in a replay buffer ${D}$ which stores joint observation every 10 episodes during learning without deletion:
\begin{equation}
\alpha=\left(\sqrt{\sum_{o_{\mathcal{N}} \in \mathcal{N}_{k}(\boldsymbol{o})} K\left(\boldsymbol{o}, \boldsymbol{o}_{\mathcal{N}}\right)}+\epsilon\right)^{-1},
\end{equation}
where $\mathcal{N}_{k}(\vo)$ is the set of $k$ nearest neighbors of $\vo$ in the replay buffer, and $\epsilon$ is a small constant (0.001) for numeric stability. Following \citet{badia2020never}, the kernel function $K(\cdot,\cdot)$ is defined as
\begin{align}
K(x, y)=\frac{\epsilon}{d^{2}(x, y)+\epsilon},
\end{align}
where $d$ is the Euclidean distance. Since the replay buffer can be large, we randomly select 10 samples from the buffer ${D}$ to approximate the $k$ nearest neighbors.

\subsection{Learning Objective}

In this paper, we use the CDS intrinsic reward~\cite{li2021celebrating} as $r^I$ and add the scaled intrinsic reward $r^\mathtt{scale}$ to environmental rewards $r^e$ for training. Furthermore, we use $Q_{\lambda}$~\cite{munos2016q, yao2021smix} for optimization. The loss function is:
\begin{equation}
\mathcal{L}_{Q_{\lambda}}(\theta)=\frac{1}{Tb}\sum_{b} \sum_{t \in\{1, \cdots, T\}}\left[G_{t}^{\lambda}-Q_{t o t}\left(s_{t}, \boldsymbol{a}_{t} ; \theta\right)\right]^{2},
\end{equation}
where b is the number of sampled trajectories used during mini-batch update, $\theta$ is the parameters of a factorized Q network~\cite{lowe2017multi, foerster2018counterfactual, sunehag2018value}, $\theta^{-}$ is the parameters of the target network which are periodically copied from $\theta$ for learning stability, and
\begin{equation}
G_{t}^{\lambda} = (1-\lambda) \sum_{n=1}^{\infty} \lambda^{n-1} G_{t}^{(n)},
\end{equation}
with 
\begin{equation}
\begin{aligned} G_{t}^{(n)}= & \left(r_{t}^{\text {scale }}+r_{t}^{e}\right)+\gamma\left(r_{t+1}^{\text {scale }}+r_{t+1}^{e}\right)+\cdots \\ & +\gamma^{n} \max _{\mathbf{a}_{t+n}} Q_{t o t}\left(s_{t+n}, \mathbf{a}_{t+n} ; \theta^{-}\right).\end{aligned}
\end{equation}

In this study, we employ the QMIX structure~\citep{rashid2018qmix} for the monotonic value factorization of $Q_{t o t}$. However, this mixing network can be readily substituted with alternative value-decomposition structures.

\section{Didactic Example}~\label{sec:toy}

\begin{figure}[t]
\centering
\includegraphics[width=1.\linewidth]{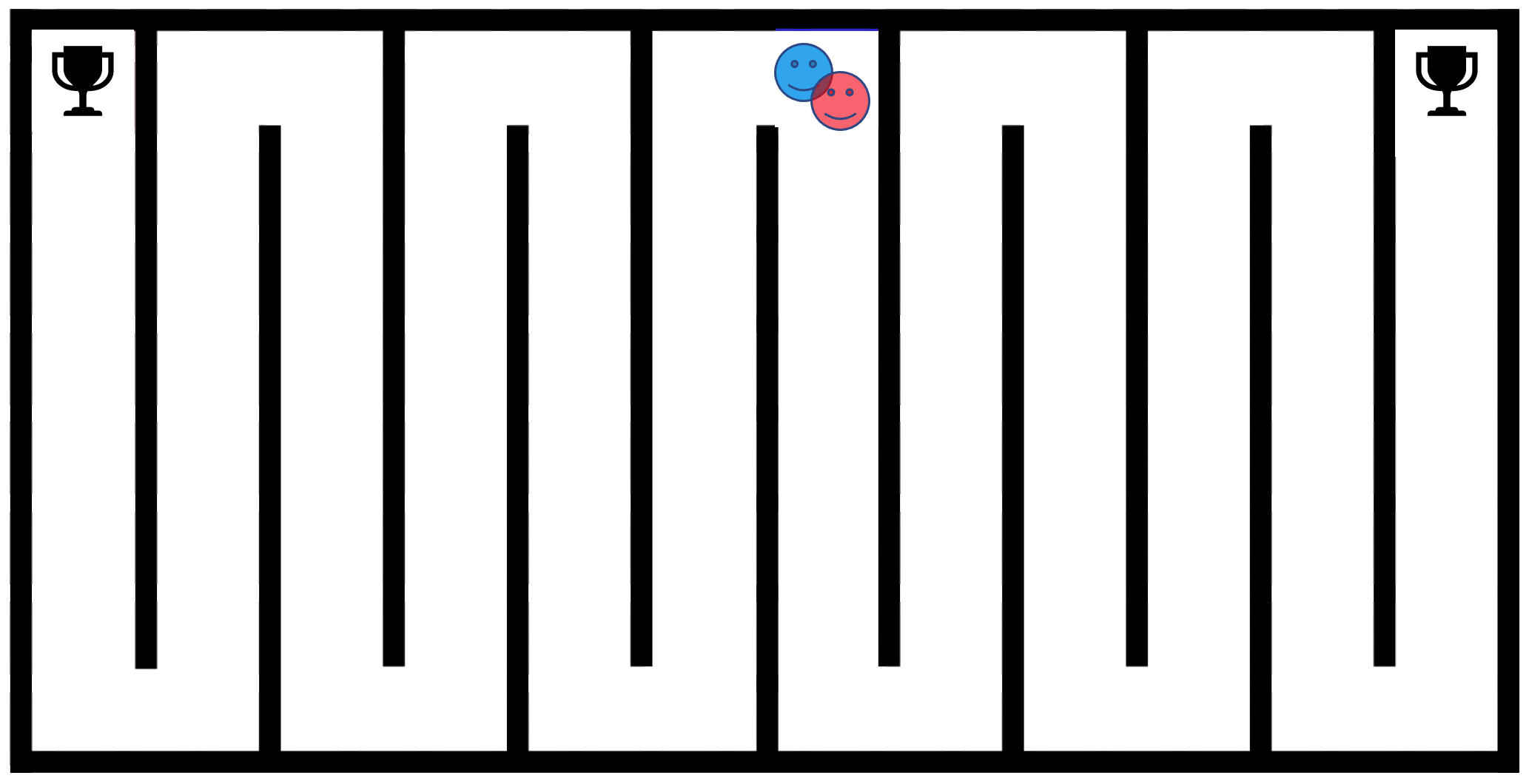}
\caption{Structure of the $\mathtt{Maze}$ task.}
\label{fig:env_maze}
\end{figure}

In this section, we introduce a maze task (as shown in Fig.~\ref{fig:env_maze}) to elucidate our concept. This task encompasses the existence of the revisitation issue, its impact on exploration, and the way our method enhances learning performance by circumventing revisitation.

In the task, two agents are placed at the center of a 6 $\times$ 12 maze and tasked with simultaneously reaching two distinct corners within 50 steps. Each agent has the ability to observe its own coordinates and those of its teammate. The action set available to both agents includes one-step movements in the four cardinal directions and an idle action that results in no movement. A team reward of 100 is bestowed only when the agents reach two different corners at the same time; otherwise, the reward is set to 0.

Fig.\ref{fig:maze_result} displays a comparison between our approach and baseline algorithms, such as the state-of-the-art value function factorization learning methods QMIX~\citep{rashid2018qmix} and QPLEX~\citep{wang2020qplex}, along with multi-agent exploration algorithms using variational approximators, such as CDS~\citep{li2021celebrating} and MAVEN~\citep{mahajan2019maven}. The experimental results highlight the superiority of our approach in exploration, as NER is the only algorithm capable of obtaining an episode reward of 100.

Furthermore, we present visitation heat maps for NER, CDS, and MAVEN across three distinct periods (0 $\sim$ 0.5M, 0.5M $\sim$ 1.0M, and 1.0M $\sim$ 1.5M). Different colors in the heat map represent different agents, with darker shades indicating higher visitation frequency. Unlike CDS and MAVEN, which show agents wandering near their initial points and wasting most training samples on revisited experiences, NER continually stimulates the agents to venture further. The propensity to re-explore previous territories by MAVEN and CDS, especially after the detachment of earlier explorations, further emphasizes the debilitating influence of revisitation on exploration. The analysis of this phenomenon illustrates that revisitation causes the exploration boundary to shrink rather than expand, limiting the agents' learning capabilities. Our approach, NER, demonstrates a practical way to overcome this limitation, enabling effective exploration that leads to better task performance.

The alterations in joint observation distributions visited by our approach are depicted in Fig.~\ref{fig:teaser} on the right side. Unlike variational intrinsic motivation methods such as MAVEN and CDS, our approach achieves sustained exploration. Specifically, for $\delta \geq 2$, the phenomenon of revisitation does not occur during the learning process. This observation underscores that, in the simple $\mathtt{Maze}$ environment, NER's dynamic adjustment of the weight of intrinsic motivations proactively prevents revisitation. In turn, this leads agents to continually extend their exploration boundaries throughout the training process. The success of NER in this context contrasts with previous methods, shedding light on a potential pathway to more efficient exploration in more complex multi-agent scenarios.

\section{Experiment}\label{sec:exp}

In the previous section, we utilized a toy maze environment to demonstrate the serious issue of repeat revisitation that plagues existing advanced multi-agent exploration algorithms. We also showcased the efficiency of our approach in facilitating exploration. In this section, we put our approach to more complex tasks, benchmarking it against standard baselines on Google Research Football (GRF) \cite{kurach2020google} and StarCraft II Micromanagement tasks (SMAC) \cite{samvelyan2019starcraft} under a sparse reward setting. These benchmarks represent some of the most challenging scenarios for cooperative multi-agent learning. We present both the median performance and the standard deviation of our method and the baselines, each tested with three different random seeds (seed=0, 1, 2). \emph{For the sake of fairness in comparison, we also apply the $Q(\lambda)$ technique in optimization to all the baseline algorithms.} 

\subsection{Google Research Football (GRF)}

\begin{figure*}[t]
\centering
\includegraphics[width=1.\linewidth]{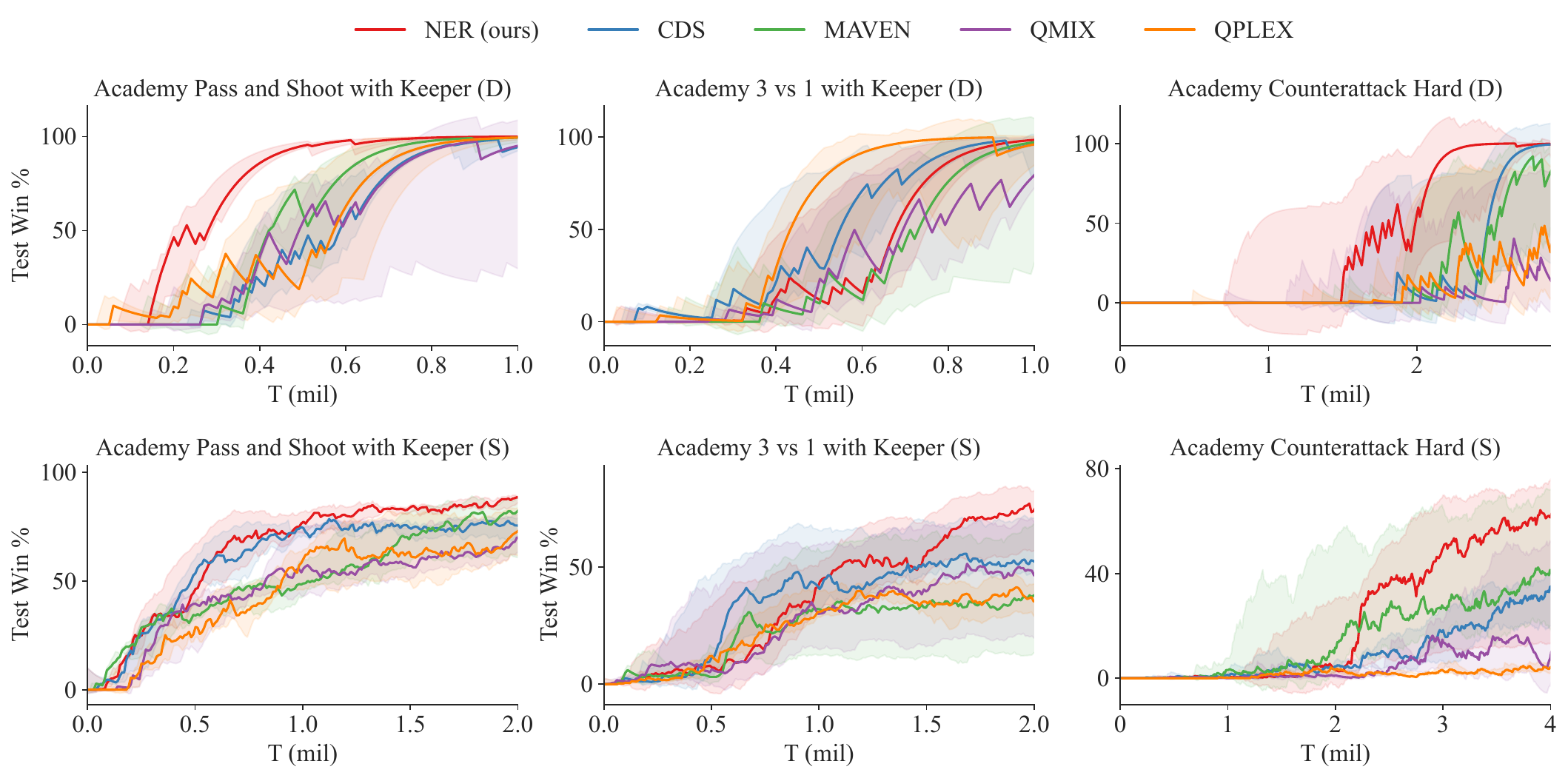}
\caption{Comparison of our approach against baseline algorithms on Google Research Football with deterministic environment seeds (D) and stochastic environment seeds (S). Using deterministic environment seeds, we evaluate the exploration capabilities of different algorithms within the environment. With stochastic environment seeds, we delve deeper to assess the exploration efficiency in relation to the diverse strategies employed by opponents.}
\label{fig:grf_deterministic}
\end{figure*}

\subsubsection{Environment Setting}

The GRF benchmark offers a variety of scenarios with differences in the number of agents and levels of difficulty. In this paper, we focus on three representative scenarios to emphasize the importance of avoiding revisitation. These scenarios include the 2-agent control scenario named $\mathtt{Academy\_Pass\_and\_Shoot\_with\_Keeper}$, $\mathtt{Academy\_3\_vs\_1\_with\_Keeper}$ with 3 controlled agents, and $\mathtt{Academy\_Counterattack\_Hard}$ with 4 controlled agents. The initial locations for agents in each of these scenarios are detailed in the Appendix.

Following the conventions established by \citet{kurach2020google} and \citet{li2021celebrating}, we use the official \emph{simple 115 representation} vector for the state and each agent's observation, constructing the observations based on relative positions. Every agent is equipped with a discrete 19-dimensional action space that encompasses various movements and basic actions like passing and shooting.

In these scenarios, environmental rewards are exclusively distributed at the episode's conclusion. Agents will receive a score of $+100$ if they manage to score, or $-1$ otherwise. For our experiments, we evaluate two different environmental random seed settings. In the deterministic setup, the random seeds are fixed, making the movement of opposing players consistent and their defensive weaknesses more exploitable. The value of deterministic environments lies in the ability to assess the exploration performance in the absence of uncertainty from opponents. It isolates the exploration behavior, enabling a more focused analysis on how agents learn to navigate and interact within the environment without the unpredictability often introduced by opponent actions. In contrast, the stochastic setting involves random seed sampling from a range of 0 to $2\times 10^9$ for each episode, leading to more unpredictable behavior from opposing players. This unpredictability increases the need for exploration, as it is more challenging to pinpoint and take advantage of defensive lapses in this more dynamic context.

\subsubsection{Experiment results}

Fig.\ref{fig:grf_deterministic} displays the comparison between our approach and various baseline algorithms in both deterministic and stochastic random seed settings. We'll first analyze the experimental results in the deterministic setting (as seen in the top portion of Fig.\ref{fig:grf_deterministic}).

In both $\mathtt{Academy\_Pass\_and\_Shoot\_with\_Keeper}$, the simplest scenario, and $\mathtt{Academy\_Counterattack\_Hard}$, the most difficult scenario, that we tested in this paper, our method consistently outperforms the baseline algorithms. The contrast between NER and CDS showcases the effectiveness of our dynamically adjusted incentive scheme in fostering exploration.

In $\mathtt{Academy\_3\_vs\_1\_with\_Keeper}$, the moderately challenging scenario, NER demonstrates rapid learning in the early stage but experiences a sudden drop in performance during the mid-stage. This occurrence reflects a classic problem in reinforcement learning: the delicate balance between exploration and exploitation. In this specific scenario, where the ball starts near the middle agent (as shown in the Appendix) and can be passed to either the top or bottom agent, different passing routes symbolize distinct cooperative strategies worth exploring. However, these diverse methods are equivalent in terms of goals, rendering continuous exploration less efficient.

Although NER eventually converges to a near 100$\%$ success rate in the $\mathtt{Academy\_3\_vs\_1\_with\_Keeper}$ scenario, the unaddressed symmetry in cooperation represents a limitation of our approach. This symmetry is challenging to detect before being explored and poses a non-trivial problem. Future work will be dedicated to investigating this issue more closely, further refining our understanding of the interaction between exploration and cooperation within multi-agent systems.

Similar to most deterministic scenarios, our approach continues to outperform the competition in the stochastic setting, as depicted in the bottom portion of Fig.~\ref{fig:grf_deterministic}. The stochastic environment introduces additional complexity due to the potentially vastly differing behaviors of opposing players across various random environment seeds. As a result, achieving a 100$\%$ average winning rate during evaluation becomes an almost unattainable goal. Despite this challenge, our approach still demonstrates clear superiority in all three scenarios tested, converging to a significantly higher winning rate. The ability of our method to adapt to the stochastic nature of the environment underscores its robustness and effectiveness. This adaptability is particularly crucial in multi-agent systems, where uncertainty and complexity are often inherent characteristics of real-world environments. Our approach's ability to excel in both deterministic and stochastic settings suggests that it could be a promising avenue for various applications where multi-agent collaboration and exploration are essential.

\subsection{StarCraft II Micromanagement (SMAC)}

\subsubsection{Environment Setting}

\begin{figure*}[ht]
\centering
\includegraphics[width=1.\linewidth]{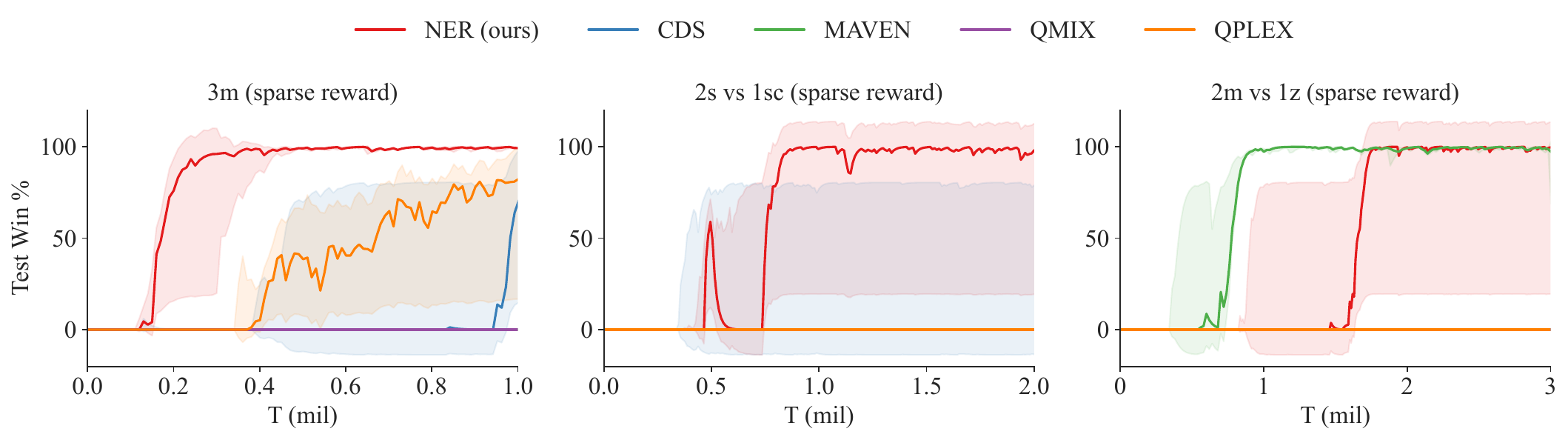}
\vspace{-2em}
\caption{Comparison of our approach against baseline algorithms on SMAC in the sparse reward setting, wherein the team receives environmental rewards exclusively at the end of episodes.}
\label{fig:smac}
\end{figure*}


In this section, we explore our approach's performance on the StarCraft II micromanagement tasks (SMAC), a complex environment comprising maps with varying numbers of agents and diverse agent types. Prior research has indeed made significant strides in many challenging SMAC tasks. However, this success is often heavily dependent on the utilization of dense rewards provided by the environment, such as the incremental changes in the health points of all agents. This reliance on dense rewards has its limitations, especially when adapting to scenarios where only sparse or terminal rewards are available.

Recognizing this challenge, we consider SMAC tasks within the sparse reward setting in this paper, which fundamentally changes the exploration landscape. Unlike the dense reward setting, where agents receive continuous feedback about their actions, the sparse reward setting only provides feedback at the end of each episode. Specifically, \emph{agents will only receive rewards at the end of one episode. They will get +100 if win, else get 0.} This setting significantly amplifies the difficulty of the tasks, requiring the agents to explore and cooperate without immediate feedback on their actions' effectiveness. Without incremental cues to guide them, the agents must learn to coordinate their strategies and actions based solely on the sparse feedback received at the episode's end.

\subsubsection{Experiment results}

Fig.\ref{fig:smac} presents a performance comparison of our method against baseline algorithms across three SMAC maps with sparse rewards. While QMIX is recognized for its impressive performance with dense rewards~\citep{rashid2018qmix}, its vanilla version faces challenges in devising effective strategies under sparse rewards, as evident in Fig.~\ref{fig:smac}.

Utilizing QMIX as its mixing network, CDS demonstrates proficiency in the $\mathtt{3m}$ scenario, attributed to its intrinsic motivations. Yet, in the $\mathtt{2m\_vs\_1z}$ map, CDS struggles to formulate a victorious strategy. In contrast, our method, which fine-tunes the intrinsic rewards of CDS by detecting revisitation patterns, consistently surpasses CDS in all three maps. This pronounced difference underscores the detrimental effects of revisitation on exploration and highlights the efficacy of our countermeasure.

Concurrently, QPLEX, boasting a duplex dueling network architecture, excels in the $\mathtt{3m}$ map but falters in the remaining two. MAVEN excels in $\mathtt{2m\_vs\_1z}$ but mirrors QMIX's performance in the other scenarios.

Our method, \name, stands out for its versatility across all maps and in other benchmarks like GRF, as previously mentioned. This showcases the broad applicability and potency of our approach, addressing not just the sparse reward challenge but also the revisitation conundrum. The superior performance of \name~underscores the transformative potential of dynamic reward adjustments, positioning it as a promising frontier for further exploration in multi-agent learning contexts.

\section{Closing Remarks}

In this paper, we delve into the subject of intrinsic motivation for multi-agent exploration, identifying and addressing the revisitation issue that hinders existing methods from realizing efficient exploration in complex tasks. By introducing a dynamic scaling scheme for intrinsic motivations, our approach not only facilitates more effective exploration but also advances the state-of-the-art performance on both GRF and SMAC benchmarks under sparse reward settings. We hope our work can encourage more studies on the limitation of current exploration methods based on variational approximations, fostering a more nuanced understanding and application of exploration strategies in multi-agent systems.

\bibliography{aaai24}

\begin{thebibliography}{59}
\providecommand{\natexlab}[1]{#1}

\bibitem[{Alemi et~al.(2017)Alemi, Fischer, Dillon, and Murphy}]{alemi2017deep}
Alemi, A.~A.; Fischer, I.; Dillon, J.~V.; and Murphy, K. 2017.
\newblock Deep Variational Information Bottleneck.
\newblock In \emph{Proceedings of the International Conference on Learning
  Representations (ICLR)}.

\bibitem[{Badia et~al.(2020)Badia, Sprechmann, Vitvitskyi, Guo, Piot,
  Kapturowski, Tieleman, Arjovsky, Pritzel, Bolt et~al.}]{badia2020never}
Badia, A.~P.; Sprechmann, P.; Vitvitskyi, A.; Guo, D.; Piot, B.; Kapturowski,
  S.; Tieleman, O.; Arjovsky, M.; Pritzel, A.; Bolt, A.; et~al. 2020.
\newblock Never give up: Learning directed exploration strategies.
\newblock \emph{arXiv preprint arXiv:2002.06038}.

\bibitem[{Bard et~al.(2020)Bard, Foerster, Chandar, Burch, Lanctot, Song,
  Parisotto, Dumoulin, Moitra, Hughes et~al.}]{bard2020hanabi}
Bard, N.; Foerster, J.~N.; Chandar, S.; Burch, N.; Lanctot, M.; Song, H.~F.;
  Parisotto, E.; Dumoulin, V.; Moitra, S.; Hughes, E.; et~al. 2020.
\newblock The hanabi challenge: A new frontier for ai research.
\newblock \emph{Artificial Intelligence}, 280: 103216.

\bibitem[{Bellemare et~al.(2016)Bellemare, Srinivasan, Ostrovski, Schaul,
  Saxton, and Munos}]{bellemare2016unifying}
Bellemare, M.; Srinivasan, S.; Ostrovski, G.; Schaul, T.; Saxton, D.; and
  Munos, R. 2016.
\newblock Unifying count-based exploration and intrinsic motivation.
\newblock In \emph{Advances in Neural Information Processing Systems},
  1471--1479.

\bibitem[{Burda et~al.(2018{\natexlab{a}})Burda, Edwards, Pathak, Storkey,
  Darrell, and Efros}]{burda2018large}
Burda, Y.; Edwards, H.; Pathak, D.; Storkey, A.; Darrell, T.; and Efros, A.~A.
  2018{\natexlab{a}}.
\newblock Large-scale study of curiosity-driven learning.
\newblock \emph{arXiv preprint arXiv:1808.04355}.

\bibitem[{Burda et~al.(2018{\natexlab{b}})Burda, Edwards, Storkey, and
  Klimov}]{burda2018exploration}
Burda, Y.; Edwards, H.; Storkey, A.; and Klimov, O. 2018{\natexlab{b}}.
\newblock Exploration by random network distillation.
\newblock \emph{arXiv preprint arXiv:1810.12894}.

\bibitem[{Chen et~al.(2021)Chen, Qiu, Ou, An, and Tambe}]{chen2021contingency}
Chen, H.; Qiu, W.; Ou, H.-C.; An, B.; and Tambe, M. 2021.
\newblock Contingency-aware influence maximization: A reinforcement learning
  approach.
\newblock In \emph{Uncertainty in Artificial Intelligence}, 1535--1545. PMLR.

\bibitem[{Ciosek et~al.(2019)Ciosek, Fortuin, Tomioka, Hofmann, and
  Turner}]{ciosek2019conservative}
Ciosek, K.; Fortuin, V.; Tomioka, R.; Hofmann, K.; and Turner, R. 2019.
\newblock Conservative uncertainty estimation by fitting prior networks.
\newblock In \emph{International Conference on Learning Representations}.

\bibitem[{de~Witt et~al.(2020)de~Witt, Gupta, Makoviichuk, Makoviychuk, Torr,
  Sun, and Whiteson}]{de2020independent}
de~Witt, C.~S.; Gupta, T.; Makoviichuk, D.; Makoviychuk, V.; Torr, P.~H.; Sun,
  M.; and Whiteson, S. 2020.
\newblock Is independent learning all you need in the starcraft multi-agent
  challenge?
\newblock \emph{arXiv preprint arXiv:2011.09533}.

\bibitem[{Dean, Tulsiani, and Gupta(2020)}]{dean2020see}
Dean, V.; Tulsiani, S.; and Gupta, A. 2020.
\newblock See, hear, explore: Curiosity via audio-visual association.
\newblock \emph{Advances in Neural Information Processing Systems}, 33:
  14961--14972.

\bibitem[{Ecoffet et~al.(2019)Ecoffet, Huizinga, Lehman, Stanley, and
  Clune}]{ecoffet2019go}
Ecoffet, A.; Huizinga, J.; Lehman, J.; Stanley, K.~O.; and Clune, J. 2019.
\newblock Go-explore: a new approach for hard-exploration problems.
\newblock \emph{arXiv preprint arXiv:1901.10995}.

\bibitem[{Endres and Schindelin(2003)}]{endres2003new}
Endres, D.~M.; and Schindelin, J.~E. 2003.
\newblock A new metric for probability distributions.
\newblock \emph{IEEE Transactions on Information theory}, 49(7): 1858--1860.

\bibitem[{Foerster et~al.(2019)Foerster, Song, Hughes, Burch, Dunning,
  Whiteson, Botvinick, and Bowling}]{foerster2019bayesian}
Foerster, J.; Song, F.; Hughes, E.; Burch, N.; Dunning, I.; Whiteson, S.;
  Botvinick, M.; and Bowling, M. 2019.
\newblock Bayesian action decoder for deep multi-agent reinforcement learning.
\newblock In \emph{International Conference on Machine Learning}, 1942--1951.

\bibitem[{Foerster et~al.(2018)Foerster, Farquhar, Afouras, Nardelli, and
  Whiteson}]{foerster2018counterfactual}
Foerster, J.~N.; Farquhar, G.; Afouras, T.; Nardelli, N.; and Whiteson, S.
  2018.
\newblock Counterfactual multi-agent policy gradients.
\newblock In \emph{Thirty-Second AAAI Conference on Artificial Intelligence}.

\bibitem[{Gupta et~al.(2021)Gupta, Mahajan, Peng, B{\"o}hmer, and
  Whiteson}]{gupta2021uneven}
Gupta, T.; Mahajan, A.; Peng, B.; B{\"o}hmer, W.; and Whiteson, S. 2021.
\newblock Uneven: Universal value exploration for multi-agent reinforcement
  learning.
\newblock In \emph{International Conference on Machine Learning}, 3930--3941.
  PMLR.

\bibitem[{Houthooft et~al.(2016)Houthooft, Chen, Duan, Schulman, De~Turck, and
  Abbeel}]{houthooft2016vime}
Houthooft, R.; Chen, X.; Duan, Y.; Schulman, J.; De~Turck, F.; and Abbeel, P.
  2016.
\newblock Vime: Variational information maximizing exploration.
\newblock In \emph{Advances in Neural Information Processing Systems},
  1109--1117.

\bibitem[{Isele and Cosgun(2018)}]{isele2018selective}
Isele, D.; and Cosgun, A. 2018.
\newblock Selective experience replay for lifelong learning.
\newblock In \emph{Proceedings of the AAAI Conference on Artificial
  Intelligence}, volume~32.

\bibitem[{Jiang and Lu(2021)}]{jiang2021emergence}
Jiang, J.; and Lu, Z. 2021.
\newblock The emergence of individuality.
\newblock In \emph{International Conference on Machine Learning}, 4992--5001.
  PMLR.

\bibitem[{Kirkpatrick et~al.(2017)Kirkpatrick, Pascanu, Rabinowitz, Veness,
  Desjardins, Rusu, Milan, Quan, Ramalho, Grabska-Barwinska
  et~al.}]{kirkpatrick2017overcoming}
Kirkpatrick, J.; Pascanu, R.; Rabinowitz, N.; Veness, J.; Desjardins, G.; Rusu,
  A.~A.; Milan, K.; Quan, J.; Ramalho, T.; Grabska-Barwinska, A.; et~al. 2017.
\newblock Overcoming catastrophic forgetting in neural networks.
\newblock \emph{Proceedings of the national academy of sciences}, 114(13):
  3521--3526.

\bibitem[{Kurach et~al.(2020)Kurach, Raichuk, Sta{\'n}czyk, Zajac, Bachem,
  Espeholt, Riquelme, Vincent, Michalski, Bousquet et~al.}]{kurach2020google}
Kurach, K.; Raichuk, A.; Sta{\'n}czyk, P.; Zajac, M.; Bachem, O.; Espeholt, L.;
  Riquelme, C.; Vincent, D.; Michalski, M.; Bousquet, O.; et~al. 2020.
\newblock Google research football: A novel reinforcement learning environment.
\newblock In \emph{Proceedings of the AAAI Conference on Artificial
  Intelligence}, volume~34, 4501--4510.

\bibitem[{Kurin et~al.(2020)Kurin, Igl, Rockt{\"a}schel, Boehmer, and
  Whiteson}]{kurin2020my}
Kurin, V.; Igl, M.; Rockt{\"a}schel, T.; Boehmer, W.; and Whiteson, S. 2020.
\newblock My Body is a Cage: the Role of Morphology in Graph-Based Incompatible
  Control.
\newblock In \emph{International Conference on Learning Representations}.

\bibitem[{Li et~al.(2021)Li, Wang, Wu, Zhao, Yang, and
  Zhang}]{li2021celebrating}
Li, C.; Wang, T.; Wu, C.; Zhao, Q.; Yang, J.; and Zhang, C. 2021.
\newblock Celebrating diversity in shared multi-agent reinforcement learning.
\newblock \emph{Advances in Neural Information Processing Systems}, 34:
  3991--4002.

\bibitem[{Li and Hoiem(2017)}]{li2017learning}
Li, Z.; and Hoiem, D. 2017.
\newblock Learning without forgetting.
\newblock \emph{IEEE transactions on pattern analysis and machine
  intelligence}, 40(12): 2935--2947.

\bibitem[{Liu et~al.(2021)Liu, Jain, Yeh, and Schwing}]{liu2021cooperative}
Liu, I.-J.; Jain, U.; Yeh, R.~A.; and Schwing, A. 2021.
\newblock Cooperative exploration for multi-agent deep reinforcement learning.
\newblock In \emph{International Conference on Machine Learning}, 6826--6836.
  PMLR.

\bibitem[{Lowe et~al.(2017)Lowe, Wu, Tamar, Harb, Abbeel, and
  Mordatch}]{lowe2017multi}
Lowe, R.; Wu, Y.; Tamar, A.; Harb, J.; Abbeel, O.~P.; and Mordatch, I. 2017.
\newblock Multi-agent actor-critic for mixed cooperative-competitive
  environments.
\newblock In \emph{Advances in Neural Information Processing Systems},
  6379--6390.

\bibitem[{Machado, Bellemare, and Bowling(2020)}]{machado2020count}
Machado, M.~C.; Bellemare, M.~G.; and Bowling, M. 2020.
\newblock Count-based exploration with the successor representation.
\newblock In \emph{Proceedings of the AAAI Conference on Artificial
  Intelligence}, volume~34, 5125--5133.

\bibitem[{Mahajan et~al.(2019)Mahajan, Rashid, Samvelyan, and
  Whiteson}]{mahajan2019maven}
Mahajan, A.; Rashid, T.; Samvelyan, M.; and Whiteson, S. 2019.
\newblock MAVEN: Multi-Agent Variational Exploration.
\newblock In \emph{Advances in Neural Information Processing Systems},
  7611--7622.

\bibitem[{Martin et~al.(2017)Martin, Sasikumar, Everitt, and
  Hutter}]{martin2017count}
Martin, J.; Sasikumar, S.~N.; Everitt, T.; and Hutter, M. 2017.
\newblock Count-based exploration in feature space for reinforcement learning.
\newblock \emph{arXiv preprint arXiv:1706.08090}.

\bibitem[{Mendez, Wang, and Eaton(2020)}]{mendez2020lifelong}
Mendez, J.; Wang, B.; and Eaton, E. 2020.
\newblock Lifelong policy gradient learning of factored policies for faster
  training without forgetting.
\newblock \emph{Advances in Neural Information Processing Systems}, 33:
  14398--14409.

\bibitem[{Munos(2016)}]{munos2016q}
Munos, R. 2016.
\newblock Q ($\lambda$) with off-policy corrections.
\newblock In \emph{Algorithmic Learning Theory: 27th International Conference,
  ALT 2016, Bari, Italy, October 19-21, 2016, Proceedings}, volume 9925, 305.
  Springer.

\bibitem[{Ndousse et~al.(2021)Ndousse, Eck, Levine, and
  Jaques}]{ndousse2021emergent}
Ndousse, K.~K.; Eck, D.; Levine, S.; and Jaques, N. 2021.
\newblock Emergent social learning via multi-agent reinforcement learning.
\newblock In \emph{International Conference on Machine Learning}, 7991--8004.
  PMLR.

\bibitem[{Oliehoek, Amato et~al.(2016)}]{oliehoek2016concise}
Oliehoek, F.~A.; Amato, C.; et~al. 2016.
\newblock \emph{A concise introduction to decentralized POMDPs}, volume~1.
\newblock Springer.

\bibitem[{Osband, Aslanides, and Cassirer(2018)}]{osband2018randomized}
Osband, I.; Aslanides, J.; and Cassirer, A. 2018.
\newblock Randomized prior functions for deep reinforcement learning.
\newblock \emph{Advances in Neural Information Processing Systems}, 31.

\bibitem[{Osband et~al.(2019)Osband, Van~Roy, Russo, Wen
  et~al.}]{osband2019deep}
Osband, I.; Van~Roy, B.; Russo, D.~J.; Wen, Z.; et~al. 2019.
\newblock Deep Exploration via Randomized Value Functions.
\newblock \emph{J. Mach. Learn. Res.}, 20(124): 1--62.

\bibitem[{Ostrovski et~al.(2017)Ostrovski, Bellemare, van~den Oord, and
  Munos}]{ostrovski2017count}
Ostrovski, G.; Bellemare, M.~G.; van~den Oord, A.; and Munos, R. 2017.
\newblock Count-based exploration with neural density models.
\newblock In \emph{Proceedings of the 34th International Conference on Machine
  Learning-Volume 70}, 2721--2730. JMLR. org.

\bibitem[{Pathak et~al.(2017)Pathak, Agrawal, Efros, and
  Darrell}]{pathak2017curiosity}
Pathak, D.; Agrawal, P.; Efros, A.~A.; and Darrell, T. 2017.
\newblock Curiosity-driven Exploration by Self-supervised Prediction.
\newblock In \emph{International Conference on Machine Learning}, 2778--2787.

\bibitem[{Pathak, Gandhi, and Gupta(2019)}]{pathak2019self}
Pathak, D.; Gandhi, D.; and Gupta, A. 2019.
\newblock Self-supervised exploration via disagreement.
\newblock In \emph{International conference on machine learning}, 5062--5071.
  PMLR.

\bibitem[{Raileanu and Rockt{\"a}schel(2020)}]{raileanu2020ride}
Raileanu, R.; and Rockt{\"a}schel, T. 2020.
\newblock Ride: Rewarding impact-driven exploration for procedurally-generated
  environments.
\newblock \emph{arXiv preprint arXiv:2002.12292}.

\bibitem[{Rashid et~al.(2018)Rashid, Samvelyan, Witt, Farquhar, Foerster, and
  Whiteson}]{rashid2018qmix}
Rashid, T.; Samvelyan, M.; Witt, C.~S.; Farquhar, G.; Foerster, J.; and
  Whiteson, S. 2018.
\newblock QMIX: Monotonic Value Function Factorisation for Deep Multi-Agent
  Reinforcement Learning.
\newblock In \emph{International Conference on Machine Learning}, 4292--4301.

\bibitem[{Rebuffi et~al.(2017)Rebuffi, Kolesnikov, Sperl, and
  Lampert}]{rebuffi2017icarl}
Rebuffi, S.-A.; Kolesnikov, A.; Sperl, G.; and Lampert, C.~H. 2017.
\newblock icarl: Incremental classifier and representation learning.
\newblock In \emph{Proceedings of the IEEE conference on Computer Vision and
  Pattern Recognition}, 2001--2010.

\bibitem[{Rolnick et~al.(2019)Rolnick, Ahuja, Schwarz, Lillicrap, and
  Wayne}]{rolnick2019experience}
Rolnick, D.; Ahuja, A.; Schwarz, J.; Lillicrap, T.; and Wayne, G. 2019.
\newblock Experience replay for continual learning.
\newblock \emph{Advances in Neural Information Processing Systems}, 32.

\bibitem[{Samvelyan et~al.(2019)Samvelyan, Rashid, de~Witt, Farquhar, Nardelli,
  Rudner, Hung, Torr, Foerster, and Whiteson}]{samvelyan2019starcraft}
Samvelyan, M.; Rashid, T.; de~Witt, C.~S.; Farquhar, G.; Nardelli, N.; Rudner,
  T.~G.; Hung, C.-M.; Torr, P.~H.; Foerster, J.; and Whiteson, S. 2019.
\newblock The starcraft multi-agent challenge.
\newblock \emph{arXiv preprint arXiv:1902.04043}.

\bibitem[{Schwarz et~al.(2018)Schwarz, Czarnecki, Luketina, Grabska-Barwinska,
  Teh, Pascanu, and Hadsell}]{schwarz2018progress}
Schwarz, J.; Czarnecki, W.; Luketina, J.; Grabska-Barwinska, A.; Teh, Y.~W.;
  Pascanu, R.; and Hadsell, R. 2018.
\newblock Progress \& compress: A scalable framework for continual learning.
\newblock In \emph{International Conference on Machine Learning}, 4528--4537.
  PMLR.

\bibitem[{Strehl and Littman(2008)}]{strehl2008analysis}
Strehl, A.~L.; and Littman, M.~L. 2008.
\newblock An analysis of model-based interval estimation for Markov decision
  processes.
\newblock \emph{Journal of Computer and System Sciences}, 74(8): 1309--1331.

\bibitem[{Sunehag et~al.(2018)Sunehag, Lever, Gruslys, Czarnecki, Zambaldi,
  Jaderberg, Lanctot, Sonnerat, Leibo, Tuyls et~al.}]{sunehag2018value}
Sunehag, P.; Lever, G.; Gruslys, A.; Czarnecki, W.~M.; Zambaldi, V.; Jaderberg,
  M.; Lanctot, M.; Sonnerat, N.; Leibo, J.~Z.; Tuyls, K.; et~al. 2018.
\newblock Value-decomposition networks for cooperative multi-agent learning
  based on team reward.
\newblock In \emph{Proceedings of the 17th International Conference on
  Autonomous Agents and MultiAgent Systems}, 2085--2087. International
  Foundation for Autonomous Agents and Multiagent Systems.

\bibitem[{Tang et~al.(2017)Tang, Houthooft, Foote, Stooke, Chen, Duan,
  Schulman, DeTurck, and Abbeel}]{tang2017exploration}
Tang, H.; Houthooft, R.; Foote, D.; Stooke, A.; Chen, O.~X.; Duan, Y.;
  Schulman, J.; DeTurck, F.; and Abbeel, P. 2017.
\newblock \# Exploration: A study of count-based exploration for deep
  reinforcement learning.
\newblock In \emph{Advances in neural information processing systems},
  2753--2762.

\bibitem[{Trott et~al.(2019)Trott, Zheng, Xiong, and Socher}]{trott2019keeping}
Trott, A.; Zheng, S.; Xiong, C.; and Socher, R. 2019.
\newblock Keeping your distance: Solving sparse reward tasks using
  self-balancing shaped rewards.
\newblock \emph{Advances in Neural Information Processing Systems}, 32.

\bibitem[{Von~Oswald et~al.(2019)Von~Oswald, Henning, Sacramento, and
  Grewe}]{von2019continual}
Von~Oswald, J.; Henning, C.; Sacramento, J.; and Grewe, B.~F. 2019.
\newblock Continual learning with hypernetworks.
\newblock \emph{arXiv preprint arXiv:1906.00695}.

\bibitem[{Wang et~al.(2021{\natexlab{a}})Wang, Ren, Liu, Yu, and
  Zhang}]{wang2020qplex}
Wang, J.; Ren, Z.; Liu, T.; Yu, Y.; and Zhang, C. 2021{\natexlab{a}}.
\newblock QPLEX: Duplex Dueling Multi-Agent Q-Learning.
\newblock \emph{International Conference on Learning Representations (ICLR)}.

\bibitem[{Wang et~al.(2020{\natexlab{a}})Wang, Dong, Lesser, and
  Zhang}]{wang2020roma}
Wang, T.; Dong, H.; Lesser, V.; and Zhang, C. 2020{\natexlab{a}}.
\newblock ROMA: Multi-Agent Reinforcement Learning with Emergent Roles.
\newblock In \emph{Proceedings of the 37th International Conference on Machine
  Learning}.

\bibitem[{Wang et~al.(2021{\natexlab{b}})Wang, Gupta, Mahajan, Peng, Whiteson,
  and Zhang}]{wang2021rode}
Wang, T.; Gupta, T.; Mahajan, A.; Peng, B.; Whiteson, S.; and Zhang, C.
  2021{\natexlab{b}}.
\newblock RODE: Learning Roles to Decompose Multi-Agent Tasks.
\newblock In \emph{Proceedings of the International Conference on Learning
  Representations (ICLR)}.

\bibitem[{Wang et~al.(2020{\natexlab{b}})Wang, Wang, Yi, and
  Zhang}]{wang2020influence}
Wang, T.; Wang, J.; Yi, W.; and Zhang, C. 2020{\natexlab{b}}.
\newblock Influence-Based Multi-Agent Exploration.
\newblock In \emph{Proceedings of the International Conference on Learning
  Representations (ICLR)}.

\bibitem[{Yan et~al.(2022)Yan, Gong, Liu, van~den Hengel, and
  Shi}]{yan2022learning}
Yan, Q.; Gong, D.; Liu, Y.; van~den Hengel, A.; and Shi, J.~Q. 2022.
\newblock Learning Bayesian Sparse Networks with Full Experience Replay for
  Continual Learning.
\newblock In \emph{Proceedings of the IEEE/CVF Conference on Computer Vision
  and Pattern Recognition}, 109--118.

\bibitem[{Yao et~al.(2021)Yao, Wen, Wang, and Tan}]{yao2021smix}
Yao, X.; Wen, C.; Wang, Y.; and Tan, X. 2021.
\newblock Smix ($\lambda$): Enhancing centralized value functions for
  cooperative multiagent reinforcement learning.
\newblock \emph{IEEE Transactions on Neural Networks and Learning Systems}.

\bibitem[{Yu et~al.(2021)Yu, Velu, Vinitsky, Wang, Bayen, and
  Wu}]{yu2021surprising}
Yu, C.; Velu, A.; Vinitsky, E.; Wang, Y.; Bayen, A.; and Wu, Y. 2021.
\newblock The surprising effectiveness of ppo in cooperative, multi-agent
  games.
\newblock \emph{arXiv preprint arXiv:2103.01955}.

\bibitem[{Zhang and Lesser(2013)}]{zhang2013coordinating}
Zhang, C.; and Lesser, V. 2013.
\newblock Coordinating multi-agent reinforcement learning with limited
  communication.
\newblock In \emph{Proceedings of the 2013 international conference on
  Autonomous agents and multi-agent systems}, 1101--1108. International
  Foundation for Autonomous Agents and Multiagent Systems.

\bibitem[{Zhang, Yang, and Zha(2019)}]{zhang2019integrating}
Zhang, Z.; Yang, J.; and Zha, H. 2019.
\newblock Integrating independent and centralized multi-agent reinforcement
  learning for traffic signal network optimization.
\newblock \emph{arXiv preprint arXiv:1909.10651}.

\bibitem[{Zheng et~al.(2021)Zheng, Chen, Wang, He, Hu, Chen, Fan, Gao, and
  Zhang}]{zheng2021episodic}
Zheng, L.; Chen, J.; Wang, J.; He, J.; Hu, Y.; Chen, Y.; Fan, C.; Gao, Y.; and
  Zhang, C. 2021.
\newblock Episodic multi-agent reinforcement learning with curiosity-driven
  exploration.
\newblock \emph{Advances in Neural Information Processing Systems}, 34:
  3757--3769.

\bibitem[{Zintgraf et~al.(2021)Zintgraf, Feng, Lu, Igl, Hartikainen, Hofmann,
  and Whiteson}]{zintgraf2021exploration}
Zintgraf, L.~M.; Feng, L.; Lu, C.; Igl, M.; Hartikainen, K.; Hofmann, K.; and
  Whiteson, S. 2021.
\newblock Exploration in approximate hyper-state space for meta reinforcement
  learning.
\newblock In \emph{International Conference on Machine Learning}, 12991--13001.
  PMLR.

\end{thebibliography}
\begin{figure*}[!htbp]
\centering
\includegraphics[width=1.\linewidth]{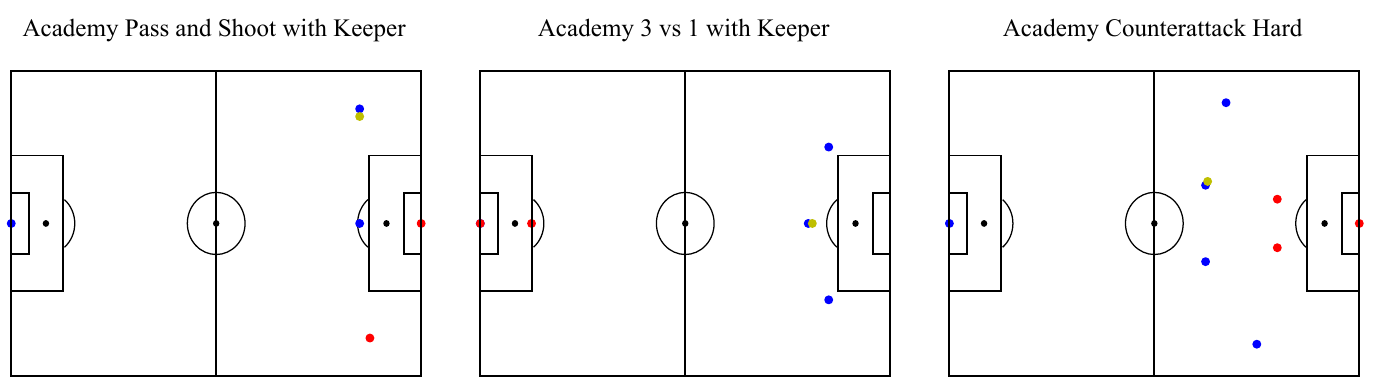}
\caption{Visualization of the initial position of each agent in three GRF scenarios considered in our paper, where blue points represent our controlled team, red points represent the opposing team, and yellow point represents the ball.}
\label{fig:position}
\end{figure*}

\newpage

\section{Appendix}
\subsection{Architecture and Hyper-parameters}

In this paper, we use simple network structures for the local Q-networks and the mixing networks as baselines. Agents use a partially shared module to represent local Q-functions as CDS~\citep{li2021celebrating}. Following the setting of CDS, agents share a trajectory encoding network made up of two layers: a fully connected layer followed by a GRU layer with a 64-dimensional hidden state for individual Q-functions. Following the trajectory encoding network, all agents share a one-layer Q network, with each agent having its own Q network with the same structure as the shared Q network. To estimate the global action values, we use QMIX-style mixing networks, which include two 32-dimensional hidden layers activated by ReLU. Hypernetworks condition on global states to generate mixing network parameters. These options apply to QMIX~\cite{rashid2018qmix} as well.

All experiments are optimized using RMSprop with a learning rate of $5 \times 10^{-4}$, $\gamma$ of 0.99, and no momentum or weight decay. We use $\epsilon$-greedy with $\epsilon$ anneals linearly from 1.0 to 0.05 over 50K time steps and keep constant for the rest of the training while selecting actions. We will evaluate the latest policies 32 times every 10k steps to calculate the average winning rate. For our approach, all baselines and ablations, we introduce a prioritized replay buffer with the same hyper-parameter of 0.3 and use TD(0.8) while calculating target values.

The intrinsic motivation introduced in CDS includes three hyper-parameters: $\beta$, $\beta_1$, and $\beta_2$, with another hyper-parameter $\lambda$ controls the weight of L1 regularization on independent local Q-networks. In this paper, we roughly fine-tune these hyper-parameters while ensuring CDS uses the same hyper-parameters during comparison. For the maze game, we use $\{\beta,\beta_1,\beta_2\, \lambda\}$ as $\{0.15,0.5,0.5,0.01\}$. For all GRF scenarios, we use $\{\beta,\beta_1,\beta_2,\lambda\}$ as $\{0.05,0.5,1.0,0.1\}$, which is the same as the original CDS setting in GRF environments. For all SMAC maps, we use $\{\beta,\beta_1,\beta_2,\lambda\}$ as $\{0.1,2.0,0.5,0.1\}$.

In our approach, the hyper-parameter $k$ is set to 10, which is the number of considered nearest neighbors.


\subsection{Visualization of GRF Scenarios}
\label{sec:setting}

The visualization of the initial position of each agent is shown in Fig.~\ref{fig:position}. The difficulty of scenarios increases from $\mathtt{Academy\_Pass\_and\_Shoot\_with\_Keeper}$ to $\mathtt{Academy\_Counterattack\_Hard}$ based on a noticeable difference between the number of agents and the initial distance of each agent to the goal and opposing players.

\end{document}